\documentclass[sigconf]{acmart}

\setcopyright{none}
\renewcommand\footnotetextcopyrightpermission[1]{}
\acmConference{}{}{}
\acmBooktitle{}

\usepackage{booktabs}
\usepackage{microtype}
\usepackage{graphicx}
\usepackage{xcolor}
\usepackage{stfloats}
\usepackage{float}
\usepackage{tikz}
\usetikzlibrary{arrows.meta, positioning, decorations.pathreplacing, calc, fit}
\usepackage{fancyhdr}

\author{Guruprasad Raghavan}
\affiliation{%
  \institution{Workfabric AI}
  \country{}
}
\email{guru@workfabric.com}

\author{George Nychis}
\affiliation{%
  \institution{Workfabric AI}
  \country{}
}
\email{george@workfabric.com}

\author{Rohan Murty}
\affiliation{%
  \institution{Workfabric AI}
  \country{}
}
\email{rohan@workfabric.com}

\begin{document}

\title{X-SYNTH: Beyond Retrieval --- Enterprise Context Synthesis from Observed Digital Human Attention}

\begin{abstract}
In enterprise operations, the context required to complete an AI agent task is scattered across systems of record, static information stores, and communication channels. What is stored is only system state, often a lossy representation of the work that actually happened~\cite{entrophy2025,tribescope2025}. The prevailing approach~\cite{lewis2020rag,guu2020realm,karpukhin2020dpr,khattab2020colbert} retrieves by matching the content of a request to what is stored; for narrow, content-driven requests such as ``Find invoice INV-34231,'' this works reasonably well. But synthesis quality depends on knowing what to surface and how to interpret it: knowledge specific to each organization, team, and individual~\cite{teevan2005personalizing,bennett2012modeling,white2009predicting}, present in behavioral patterns, absent from any retrieval index or general training corpus. For complex agentic tasks requiring synthesis from dispersed behavioral evidence, it breaks down: True Lead Rate is low, False Lead Rate is high, and the model has no mechanism to improve.

We present \textbf{X-SYNTH}, a framework for enterprise context synthesis grounded in digital human attention — the digitally observable interaction signatures of each worker, encoding not just what they did but the sequence in which they did it. Enterprise workers continuously produce digital interaction data~\cite{gonzalez2004multitasking,mark2005notask,czerwinski2004diary,entrophy2025,tribescope2025} that encodes not just outcomes but the ordered sequence of actions by which they were reached, along with the implicit reward signals embedded within them~\cite{joachims2002clickthrough,joachims2005accurately,kelly2003implicit,agichtein2006improving,buscher2008eye}. Everything needed to learn relevance is present: behavioral traces that preceded positive outcomes are distinguishable from those that did not, without any external labeling. X-SYNTH models each individual's behavioral baseline as a Digital Twin Signature (DTS) and selects among seven qualitatively distinct attention filters: Proportional, Inverse, Differential, Recurrent, Comparative, Sequential, and Collective, per individual and per query, to identify which activity signatures were causally relevant. A four-stage agentic pipeline assembles this into ranked context grounded in behavioral patterns rather than query embeddings.

On a sales lead identification task, a frontier model unaided achieves a 9.5\% True Lead Rate (TLR) with a 90.5\% False Lead Rate (FLR). Augmented with X-SYNTH, TLR rises to 61.9\% (a 6.5$\times$ improvement) while FLR falls to 18.8\%. Enterprise context synthesis is not a retrieval problem. It is a relevance problem, and digital human attention is its most reliable ground truth.

\end{abstract}



\maketitle

\pagestyle{fancy}
\fancyhf{}
\renewcommand{\headrulewidth}{0pt}
\setlength{\footskip}{40pt}
\cfoot{\thepage}

\section{Introduction}
\label{sec:introduction}

Consider a sales representative closing a deal that has been in motion for three months. The outcome, a signed contract, will eventually appear in a CRM. But the context that produced it will not: the email thread that resolved a pricing concern, the Slack message that unblocked a technical objection, the internal document revised four times before it moved. None of these artifacts were written to be retrieved together. What is relevant to closing the next deal, or finding a new one altogether? And when an AI agent is asked to do exactly that, none of them will surface, because no retrieval system was watching when they mattered.

Without solving for this level of complexity, AI agents in enterprise operations remain largely limited to fetching static information and executing explicitly defined rules on how to act on it. Ask an agent~\cite{yao2023react,schick2023toolformer} to retrieve invoice \#12345 and it will succeed. Ask it to find a new sales lead for your team and it will not~\cite{herb2025}, because finding a new lead is not a retrieval task: there is no record of it waiting to be found. The agent must \emph{synthesize one} from the behavioral evidence embedded in the team's history of activity, and the prevailing approach of connecting systems of record and retrieving against the content of a request is not designed for this.

The consequences are measurable, and recent enterprise-specific benchmarks confirm them at scale~\cite{herb2025,enterprisebench2025,wornow2024automating,muthusamy2023personal,marreed2025enterprise}. In one of our benchmarks, a sales lead identification task in which an agent must surface net-new opportunities from a seller's digital interaction history before they are recorded in a CRM, Claude Opus 4.6 produces a \textbf{90.5\% False Lead Rate (FLR)} (only 1 in 10 surfaced leads is real) and misses \textbf{90.5\% of actual leads}, catching just 20 of the 210 positive instances in the benchmark. The model's intelligence is not the bottleneck; the absence of a principled relevance signal towards outcomes such as this is.

Digital human attention — the ordered, digitally observable signature of who focuses on what, in what sequence, and how that deviates from their behavioral baseline — is the ground truth that retrieval-based approaches lack. Enterprise workers continuously produce digital interaction data~\cite{gonzalez2004multitasking,mark2005notask,czerwinski2004diary,entrophy2025,tribescope2025} that encodes not just outcomes but the ordered sequence of actions by which they were reached. Knowing what signals matter and how to interpret them is specific to each organization, team, geography, and individual. No general training corpus encodes that a developer's absence from security reviews is anomalous for \emph{this person specifically}, or that an account executive's attention shifting to competitor documents signals deal risk; that knowledge lives in the organization's own behavioral history.

X-SYNTH exploits this structure through seven qualitatively distinct attention filters: Proportional, Inverse, Differential, Recurrent, Comparative, Sequential, and Collective, each designed to surface a different behavioral signal. The same query resolves to a different filter for each individual: a developer who owns security tooling warrants a Proportional filter on their security artifact attention, while one who has gone unusually quiet warrants a Differential filter. Selection is conditioned on each individual's Digital Twin Signature (DTS), a compact rolling behavioral profile, via a learned function \emph{Query}~$\times$~\emph{DTS}~$\rightarrow$~\emph{Modality}. A four-stage agentic pipeline assembles the result: subject scoping, individualized modality selection, attention-and-content-weighted retrieval, and synthesis.

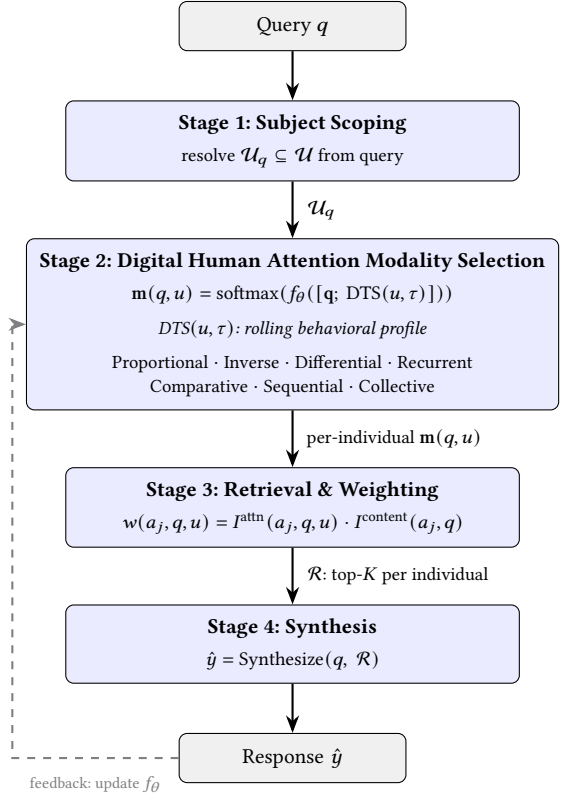
\begin{figure}[H]
\centering
\begin{tikzpicture}[
  node distance=0.75cm,
  every node/.style={font=\small},
  stage/.style={draw, rounded corners=3pt, fill=blue!8, minimum width=6.0cm, align=center, inner sep=5pt},
  io/.style={draw, rounded corners=3pt, fill=gray!12, minimum width=3cm, minimum height=0.65cm, align=center},
  arrow/.style={-Stealth, thick},
]

\node[io] (query) {Query $q$};

\node[stage, below=0.65cm of query] (stage1)
  {\textbf{Stage 1: Subject Scoping}\\[2pt]
   \footnotesize{resolve $\mathcal{U}_q \subseteq \mathcal{U}$ from query}};

\node[stage, below=0.75cm of stage1] (stage2)
  {\textbf{Stage 2: Digital Human Attention Modality Selection}\\[2pt]
   \footnotesize{$\mathbf{m}(q,u) = \mathrm{softmax}(f_\theta([\mathbf{q};\ \mathrm{DTS}(u,\tau)]))$}\\[3pt]
   \footnotesize{\textit{DTS$(u,\tau)$: rolling behavioral profile}}\\[3pt]
   \footnotesize{Proportional $\cdot$ Inverse $\cdot$ Differential $\cdot$ Recurrent}\\[-1pt]
   \footnotesize{Comparative $\cdot$ Sequential $\cdot$ Collective}};

\node[stage, below=0.75cm of stage2] (stage3)
  {\textbf{Stage 3: Retrieval \& Weighting}\\[2pt]
   \footnotesize{$w(a_j,q,u) = I^{\mathrm{attn}}(a_j,q,u)\cdot I^{\mathrm{content}}(a_j,q)$}};

\node[stage, below=0.75cm of stage3] (stage4)
  {\textbf{Stage 4: Synthesis}\\[2pt]
   \footnotesize{$\hat{y} = \mathrm{Synthesize}(q,\ \mathcal{R})$}};

\node[io, below=0.65cm of stage4] (output) {Response $\hat{y}$};

\draw[arrow] (query)  -- (stage1);
\draw[arrow] (stage1) -- node[right, font=\footnotesize, xshift=2pt] {$\mathcal{U}_q$} (stage2);
\draw[arrow] (stage2) -- node[right, font=\footnotesize, xshift=2pt, align=left]
                           {per-individual $\mathbf{m}(q,u)$} (stage3);
\draw[arrow] (stage3) -- node[right, font=\footnotesize, xshift=2pt, align=left]
                           {$\mathcal{R}$: top-$K$ per individual} (stage4);
\draw[arrow] (stage4) -- (output);

\draw[-Stealth, thick, dashed, gray]
  (output.west) -- ++(-2.2,0) coordinate (fb1)
  -- (fb1 |- stage2.west) coordinate (fb2)
  -- (stage2.west);
\node[below=0.12cm, font=\scriptsize, text=gray, align=center]
  at ($(output.west)!0.5!(fb1)$) {feedback: update $f_\theta$};

\end{tikzpicture}
\caption{The X-SYNTH pipeline. A query is scoped to target individuals $\mathcal{U}_q$; each is assigned a modality distribution over seven attention filters conditioned on their Digital Twin Signature (DTS) and the query. Artifacts are ranked by combined attention-and-content importance; a synthesis stage assembles the response. The same query resolves to different filters for different individuals.}
\label{fig:pipeline}
\end{figure}

X-SYNTH makes the following contributions:

\begin{itemize}
  \item \textbf{Digital human attention as behavioral ground truth.} Digital human attention — operationalized as the ordered, digitally observed interaction sequence of each worker — is a learnable and discriminative relevance signal without external labels. Each individual's Digital Twin Signature (DTS) encodes this as a compact rolling behavioral profile.

  \item \textbf{Seven-filter attention modality framework.} Seven qualitatively distinct attention filters (Proportional, Inverse, Differential, Recurrent, Comparative, Sequential, Collective), selected per individual and per query via a lightweight MLP conditioned on \emph{Query}~$\times$~\emph{DTS}.

  \item \textbf{Four-stage agentic synthesis pipeline.} Subject scoping, individualized modality selection, attention-and-content-weighted retrieval, and synthesis, with per-individual filter assignment grounded in each person's behavioral baseline.

  \item \textbf{Feedback loop with credit attribution.} X-SYNTH decomposes terminal feedback into stage-level failure probabilities, updating the modality selector only when failure is attributed to modality misclassification rather than retrieval or synthesis error, closing the improvement loop without polluting the training signal.
\end{itemize}

Augmenting the same Claude Opus 4.6 model with X-SYNTH yields substantial improvements. True Lead Rate (TLR) jumps from \textbf{9.5\% to 61.9\%} (a \textbf{6.5$\times$ improvement}), capturing 130 vs.\ 20 of 210 real leads, while FLR falls from \textbf{90.5\% to 18.8\%}. In absolute terms, X-SYNTH surfaces \textbf{110 additional real leads} that the out-of-the-box approach misses entirely.


\section{Related Work}
\label{sec:related-work}

X-SYNTH sits at the intersection of retrieval-augmented generation, implicit feedback in information retrieval, personalized user modeling, knowledge-worker behavioral studies, enterprise workflow capture, LLM agents, and learned routing. We survey each lineage and position X-SYNTH within it.

\subsection{Retrieval-Augmented Generation}
\label{subsec:rw-rag}

Retrieval-augmented generation~\cite{lewis2020rag,guu2020realm} grounds language model outputs in retrieved passages, with dense~\cite{karpukhin2020dpr}, sparse~\cite{robertson2009bm25}, and hybrid~\cite{khattab2020colbert} retrievers as standard components. Subsequent work has explored larger retrieval pools~\cite{borgeaud2022retro,izacard2023atlas} and architectural variants for fusing retrieved content~\cite{izacard2021fid}. A recent thread makes retrieval adaptive --- deciding whether~\cite{asai2024selfrag}, when~\cite{jiang2023flare,su2024dragin}, or how to correct~\cite{yan2024crag} mid-generation --- and black-box variants~\cite{shi2024replug,ram2023icrag} extend the paradigm to closed models. Across this literature, the relevance signal is the \textit{query}: passages are selected by content similarity, optionally filtered by self-judged sufficiency. It struggles when the answer is dispersed across artifacts that share no surface similarity to the query, as recent enterprise-specific benchmarks demonstrate~\cite{herb2025,enterprisebench2025}: HERB~\cite{herb2025} reports best-in-class agentic RAG averaging $\sim$33\% on a heterogeneous enterprise corpus and identifies retrieval as the bottleneck. X-SYNTH adds an orthogonal signal --- observed digital human attention --- rather than improving content-based retrieval.

\subsection{Implicit Feedback as Relevance Signal}
\label{subsec:rw-implicit}

The premise that user behavior carries relevance information predates the RAG era. Joachims~\cite{joachims2002clickthrough} showed that clickthrough data yields ranking signals competitive with editorial judgments; subsequent work refined the interpretation~\cite{joachims2005accurately} and addressed selection bias~\cite{joachims2017unbiased}. The signal vocabulary expanded to dwell time~\cite{claypool2001implicit}, revisits, and post-click actions~\cite{kelly2003implicit,fox2005evaluating}, with eye-tracking studies grounding these in digital human attentional mechanisms~\cite{buscher2008eye}. Agichtein et al.~\cite{agichtein2006improving} integrated behavioral and content features in a unified ranker --- a formulation X-SYNTH generalizes in \S\ref{sec:methodology}, where attention importance and content relevance combine multiplicatively. Sequential modeling~\cite{hidasi2016gru4rec,kang2018sasrec,sun2019bert4rec,zhou2018din} established that the \textit{order} of activity carries information beyond its aggregate. X-SYNTH inherits both intuitions --- behavior as implicit reward, sequence as structure --- and extends them: the behavioral substrate is enterprise activity over heterogeneous artifacts rather than search clicks; relevance is structured into seven qualitatively distinct filters; and an explicit per-individual baseline (the DTS) replaces population-averaged relevance.

\subsection{Personalized Search and User Modeling}
\label{subsec:rw-personalization}

Personalization in IR began with the recognition that the same query should yield different results for different users~\cite{teevan2005personalizing}. Building per-user profiles from prior activity enabled re-ranking without requiring users to specify intent~\cite{sugiyama2004adaptive,shen2005context,dou2007personalized}. Bennett et al.~\cite{bennett2012modeling} showed that short-term and long-term signals are complementary, motivating multi-window user models; X-SYNTH's $\mathbf{v}^{\text{div}}$ component (KL divergence between 5-day and 14-day domain attention) directly inherits this insight. Contextual user-interest prediction~\cite{white2009predicting} and personal corpus retrieval~\cite{dumais2003stuff} further established that behavioral history is searchable and predictive. The DTS differs from prior user models in that it is built on enterprise activity streams rather than search interactions, it is structured into five distinct behavioral components, and it is consumed by a learned router ($f_\theta$) rather than used directly as ranker features --- allowing the same query to resolve to different filters for different individuals.

\subsection{Knowledge-Worker Behavior and Enterprise Capture}
\label{subsec:rw-knowledge-work}

Gonz\'{a}lez and Mark~\cite{gonzalez2004multitasking} showed that information workers operate across roughly ten working spheres simultaneously, switching every twelve minutes on average; subsequent studies confirmed the pattern~\cite{mark2005notask,czerwinski2004diary,iqbal2007disruption}. Work rhythms are themselves structured and predictable~\cite{mark2014bored}, supporting the hypothesis that an individual's behavioral baseline is a meaningful, learnable object. Activity-aware systems~\cite{dragunov2005tasktracer,bardram2005activity} established that capturing activity as a first-class signal is technically feasible, and enterprise search has long been recognized as structurally distinct from web search~\cite{hawking2004enterprise}. Operationalizing behavior-as-signal at scale requires high-fidelity interaction capture~\cite{tribescope2025}; ENTROPHY~\cite{entrophy2025} provides the first large-scale dataset of real production workflows averaging 178 steps, confirming that frontier models cannot treat interaction streams as plain text. Process mining~\cite{vanderaalst2016processmining} and digital interaction intelligence~\cite{modi2024dii} share the same input but target process discovery rather than query-conditioned synthesis.

\subsection{Enterprise LLM Agents}
\label{subsec:rw-enterprise-agents}

A growing body of work explores LLMs as enterprise actors~\cite{wornow2024automating,muthusamy2023personal,rizk2024bpfm,marreed2025enterprise}, with Kayali et al.~\cite{kayali2025datagap} framing the gap between LLM capabilities and enterprise data integration. These works largely converge on a finding consistent with ours: enterprise context is the missing ingredient, and content-based retrieval alone is insufficient. X-SYNTH operationalizes one specific bridge --- observed digital human attention as the relevance ground truth.

\subsection{LLM Agents: Reasoning, Tools, and Self-Improvement}
\label{subsec:rw-agents}

The agentic harness X-SYNTH is embedded in draws on the reasoning-and-acting paradigm~\cite{yao2023react} built on chain-of-thought reasoning~\cite{wei2022cot}. Tool-using LMs~\cite{schick2023toolformer,qin2024toolllm} established that LLMs can route between specialized capabilities; multi-agent frameworks~\cite{wu2024autogen,hong2024metagpt} generalize this to coordinated specialists. Self-improvement via terminal feedback has been explored~\cite{shinn2023reflexion,madaan2023selfrefine,wang2023voyager}; the CoALA framework~\cite{sumers2024coala} organizes agents into modular memory, action, and decision components. X-SYNTH's contribution at the agent layer is the \textit{attribution} of terminal feedback to pipeline stages: rather than treating a poor response as evidence to update every component~\cite{shinn2023reflexion}, we decompose failure into stage-level probabilities and update the modality selector only when failure is attributed to Stage~2.

\subsection{Mixture-of-Experts and Learned Routing}
\label{subsec:rw-moe}

The modality selector $\mathbf{m}(q,u) = \text{softmax}(f_\theta([\mathbf{q};\text{DTS}(u,\tau)]))$ is a soft router over seven specialist filters, echoing the mixture-of-experts lineage from Jacobs et al.~\cite{jacobs1991moe} through modern sparse MoE~\cite{shazeer2017moe,fedus2022switch,lepikhin2021gshard}. The key structural difference is that the router is conditioned on the \textit{user state} (the DTS) in addition to the query, so the same query routes to different filters for different individuals --- a property that standard input-only MoE gating does not exhibit.

\section{Methodology}
\label{sec:methodology}

X-SYNTH is an agentic retrieval and synthesis system that resolves a natural language query $q$ against a corpus of human interaction traces $\mathcal{E}$. The pipeline proceeds through four stages as shown in Figure~\ref{fig:pipeline}: (i) subject scoping, (ii) attention modality selection, (iii) artifact retrieval and importance weighting, and (iv) response synthesis. Table~\ref{tab:notation} summarizes key notation.

\begin{table}[h]
\small
\centering
\begin{tabular}{ll}
\toprule
Symbol & Definition \\
\midrule
$q$ & Natural language query \\
$\mathbf{q} \in \mathbb{R}^{d_q}$ & Query embedding \\
$\mathcal{U}_q \subseteq \mathcal{U}$ & Individuals scoped by query \\
$\mathcal{A} = \{a_1, \ldots, a_M\}$ & Set of artifacts \\
$\mathcal{E}$ & Raw interaction event log \\
$\tau$ & Rolling behavioral window \\
$\text{DTS}(u, \tau)$ & Digital Twin Signature of individual $u$ \\
$\mathbf{m}(q, u) \in \Delta^6$ & Modality probability vector over 7 filters \\
$w(a_j, q, u)$ & Final importance weight of artifact $a_j$ \\
\bottomrule
\end{tabular}
\caption{Key notation used throughout \S\ref{sec:methodology}. The DTS vector $\text{DTS}(u,\tau)$ concatenates five behavioral components over a rolling behavioral window; $\mathbf{m}(q,u) \in \Delta^6$ is the learned soft assignment over the seven attention filters; $w(a_j,q,u)$ is the final artifact importance combining attention and content signals.}
\label{tab:notation}
\end{table}

\subsection{Stage 1: Subject Scoping}

Given query $q$, the system identifies the target set of individuals:
\[ \mathcal{U}_q = \text{Scope}(q, \mathcal{U}) \]

If $q$ references a named individual or group (e.g., \textit{``John''}, \textit{``the security team''}), $\mathcal{U}_q$ is resolved via entity extraction. If no subject constraint is present, $\mathcal{U}_q = \mathcal{U}$. Subject scoping is deterministic and requires no learned components.

\subsection{Stage 2: Attention Modality Selection}

\subsubsection{Digital Twin Signature}

For each individual $u \in \mathcal{U}_q$, the system maintains a Digital Twin Signature computed over a rolling behavioral window $\tau$:
\[ \text{DTS}(u, \tau) = \left[ \mathbf{v}^{\text{dom}},\ \mathbf{v}^{\text{rhythm}},\ \mathbf{v}^{\text{base}},\ \mathbf{v}^{\text{resp}},\ \mathbf{v}^{\text{div}} \right] \]

where the five components encode:

\begin{itemize}
  \item $\mathbf{v}^{\text{dom}}$ --- \textbf{domain attention}: where attention is concentrated across work domains.
  \item $\mathbf{v}^{\text{rhythm}}$ --- \textbf{behavioral rhythm}: statistics capturing dwell, revisit, and transition patterns over the window.
  \item $\mathbf{v}^{\text{base}}$ --- \textbf{baseline}: per-domain baseline statistics over an extended lookback, establishing what is normal for this individual.
  \item $\mathbf{v}^{\text{resp}}$ --- \textbf{responsibility profile}: inferred domain ownership derived from activity patterns, not declared role.
  \item $\mathbf{v}^{\text{div}}$ --- \textbf{short-vs-long divergence}: how recent attention patterns differ from the individual's longer-term baseline, per domain.
\end{itemize}

\subsubsection{The Seven Attention Filters}

X-SYNTH defines seven qualitatively distinct attention filters $\mathcal{M} = \{M_1, \ldots, M_7\}$, each designed to surface a different behavioral signal. Table~\ref{tab:filters} describes each and its intended use.

\begin{table}[h]
\small
\centering
\begin{tabular}{p{1.6cm}p{5.8cm}}
\toprule
Filter & Description \\
\midrule
Proportional & Higher dwell time signals higher importance. Used when the subject's attention is expected and meaningful for the domain. \\[2pt]
Inverse & Artifacts that \emph{should} have received attention but did not. Absence is the signal. \\[2pt]
Differential & Importance is determined by deviation from the individual's own behavioral baseline, not absolute dwell. \\[2pt]
Recurrent & Artifacts that pulled the subject back repeatedly. Return frequency, not dwell, is the signal. \\[2pt]
Comparative & Rapid alternation between semantically similar artifacts signals active evaluation. \\[2pt]
Sequential & The order of attention matters. Deviations from expected workflow sequences are flagged. \\[2pt]
Collective & Aggregates attention across a cohort to surface consensus focus and individual outliers. \\
\bottomrule
\end{tabular}
\caption{The seven attention filters of X-SYNTH, each operationalizing a qualitatively distinct relevance signal. Proportional and Recurrent filters reward high absolute attention; Inverse rewards notable absence; Differential rewards deviation from an individual's own baseline; Comparative rewards rapid alternation between similar artifacts; Sequential rewards workflow-order anomalies; Collective aggregates across a cohort. The modality selector $f_\theta$ assigns a soft distribution over these filters per $(q, u)$ pair conditioned on the query and the individual's DTS.}
\label{tab:filters}
\end{table}

\subsubsection{Modality Selection}

The central claim is that the same query resolves to a different filter for each individual in $\mathcal{U}_q$, conditioned on their behavioral baseline. For each $(q, u)$ pair, the modality selector produces a distribution over $\mathcal{M}$:

\[ \mathbf{m}(q, u) = \text{softmax}\!\left( f_\theta\!\left( [\mathbf{q};\ \text{DTS}(u, \tau)] \right) \right) \in \Delta^6 \]

where $f_\theta : \mathbb{R}^{d_q + 5d + 6} \rightarrow \mathbb{R}^7$ is a three-layer MLP with hidden dimensions 256 and 64. The output is a soft assignment; in practice, one or two filters dominate for any given $(q, u)$ pair.

To illustrate: for the query \textit{``Is the team on top of the security vulnerabilities flagged this sprint?''}, a developer who owns the security tooling and regularly reviews CVE reports triggers a Proportional filter — their attention is the signal. A developer who used to own security reviews but whose DTS shows a sharp drop in security artifact interaction this sprint triggers a Differential filter — the deviation from their own baseline is the signal. A developer who rarely touches security at all triggers an Inverse filter — their absence of engagement is the signal. Same query; three different filters; three different answers surfaced.

\subsubsection{Training}

Training follows a two-phase strategy. In \textbf{Phase 1}, a rule-based classifier $f_{\text{rule}}$ assigns modalities based on linguistic cues in the query. This covers the majority of queries where the correct modality is unambiguous, independent of the DTS.

In \textbf{Phase 2}, cases where $f_{\text{rule}}$ produces ambiguous or conflicting signals are collected and labeled by domain experts, yielding a training set $\{(\mathbf{q}_i, \text{DTS}(u_i, \tau_i), k_i^*)\}$. The MLP $f_\theta$ is trained on this harder subset via cross-entropy loss:

\[ \mathcal{L}(\theta) = -\sum_i \log m_{k_i^*}(q_i, u_i) \]

This strategy ensures $f_\theta$ allocates capacity to the cases where the DTS is genuinely discriminative: where role, behavioral baseline, or a recent shift in behavior determines the correct filter.

\subsection{Stage 3: Artifact Retrieval and Importance Weighting}

For each $u \in \mathcal{U}_q$, artifacts are ranked by a weight that combines behavioral attention signal with content relevance.

\textbf{Attention-based importance.} Each modality $M_k$ defines an importance function $I_k : \mathcal{A} \rightarrow \mathbb{R}_{\geq 0}$. Given the soft modality vector $\mathbf{m}(q, u)$, the blended attention importance is:

\[ I^{\text{attn}}(a_j, q, u) = \sum_{k=1}^{7} m_k(q, u) \cdot I_k(a_j,\ u,\ \tau) \]

where each $I_k$ scores artifacts using the individual's behavioral data under filter $M_k$.

\textbf{Content-based relevance.} Each artifact is also scored for relevance to $q$ by combining lexical and semantic signals, producing a content relevance score $I^{\text{content}}(a_j, q)$.

\textbf{Combined weight.} The final importance weight combines both signals multiplicatively:

\[ w(a_j, q, u) = I^{\text{attn}}(a_j, q, u) \cdot I^{\text{content}}(a_j, q) \]

The multiplicative formulation ensures an artifact must be both relevant to the query \emph{and} meaningfully attended to. Artifacts are ranked by $w(a_j, q, u)$ and the top-$K$ are selected per individual.

\subsection{Stage 4: Agentic Synthesis}

\textbf{Tool selection.} X-SYNTH is embedded within an agentic harness in which the attention modality selector is one of several available tools. The attention tool is invoked when the query is evidence-seeking over human behavior. When invoked, it returns for each $u \in \mathcal{U}_q$ the ranked artifact set $\{(a_j, w(a_j, q, u))\}_{j=1}^{K}$ along with modality annotations explaining why each artifact was surfaced.

\textbf{Synthesis.} Once evidence is gathered across all $u \in \mathcal{U}_q$, a synthesis step produces the response. Let $\mathcal{R}_u$ denote the retrieved evidence set for individual $u$. The synthesizer receives $\mathcal{R} = \bigcup_{u \in \mathcal{U}_q} \mathcal{R}_u$ and produces:

\[ \hat{y} = \text{Synthesize}(q,\ \mathcal{R}) \]

where $\text{Synthesize}(\cdot)$ is an LLM prompted with the query, the ranked evidence, and the modality annotations. The annotations are critical: they provide the synthesizer with an explicit account of \emph{why} each artifact was retrieved, grounding the response and reducing hallucination.

\textbf{Feedback loop.} Terminal feedback (user satisfaction signal $s \in \{0,1\}$) is insufficient for training the modality selector directly, as a poor response may stem from modality misclassification, retrieval failure, or synthesis error. X-SYNTH maintains intermediate signals to decompose feedback into stage-level attribution probabilities:

\[ P(\text{failure at stage } k \mid s=0,\ \mathcal{R},\ \hat{y}) \]

The modality selector is updated only when failure is attributed to Stage 2 with sufficient confidence, closing the improvement loop without polluting the training signal with synthesis-layer errors.

\section{Evaluation}
\label{sec:evaluation}

\subsection{Dataset}

We evaluate on a dataset of real-world digital interaction traces collected from five knowledge workers --- spanning roles of Account Executives, Sales Directors, and Client Resource Managers --- at an anonymized Fortune 500 enterprise. Data collection covered 25 working days per participant, yielding 125 seller journeys in total.

Each interaction record is a structured tuple comprising: participant ID, application accessed, UTC timestamp, screen content (title, key UI attributes, and on-screen text), user action, and dwell time. A single knowledge worker generates on the order of 1{,}000 interactions per day, with each interaction carrying up to 1{,}600 LLM tokens of context. This yields approximately 1.6M tokens per worker per day; across five workers and 25 days, the full dataset comprises roughly 200M tokens of raw interaction data. All participant data was anonymized prior to analysis in accordance with enterprise data handling agreements.

\subsection{Benchmark Construction}

The core evaluation task is: given a sequence of raw digital interactions, can a system detect and synthesize a latent business opportunity that a seller would legitimately file in their sales platform?

To construct ground-truth labels without manual annotation, we exploit a naturally occurring signal in the data: sellers routinely file new business opportunities directly within their seller platform, and these filing events appear as interactions in the raw log. Our benchmark construction proceeds as follows. We identify all instances where a seller filed a new business opportunity --- the \emph{pivot point} --- and reconstruct the preceding interaction sequence as the system input. The pivot point itself, along with any directly associated filing-screen interactions, is removed from the input; only interactions that preceded and plausibly informed the opportunity are retained --- emails received, Teams conversations, Excel updates, SOW documents reviewed in Word, and similar. The removed filing event and its structured content serve as the ground-truth target.

This yields a benchmark of \textbf{302 instances}: 210 positive instances where a genuine business opportunity was filed, and 92 negative instances where the seller completed routine work with no opportunity filed.

\subsection{Metrics}

We report three primary metrics, motivated directly by business impact:

\textbf{True Lead Rate (TLR).} The fraction of positive instances in which the system successfully surfaces a recommendation that matches the ground-truth opportunity. This is the primary measure of business value --- leads correctly identified represent revenue opportunities that the system has made actionable for the seller.

\textbf{Missed Lead Rate (MLR).} The fraction of positive instances in which the system fails to surface a recommendation corresponding to the ground-truth opportunity. A missed lead represents lost revenue potential --- a signal present in the data that the system did not act on. Note that TLR and MLR are complementary: $\text{MLR} = 1 - \text{TLR}$ over positive instances.

\textbf{False Lead Rate (FLR).} The fraction of instances in which the system recommends a business opportunity that does not correspond to a valid ground-truth filing. Each false lead imposes a direct cost on the seller, requiring manual review and discard.

\textbf{Evaluation is automated.} For each system-proposed lead, we compute embedding similarity between the proposed opportunity description and the ground-truth filing to assess semantic alignment. For interaction attribution --- whether the system correctly identified the underlying evidence that led to the opportunity --- we score based on key attribute match: a proposed evidence set receives full credit if its key attributes align with those captured in the ground-truth dataset for that instance. This attribution metric is naturally applicable to systems, like X-SYNTH, that produce structured evidence traces; for baseline models that operate over the raw interaction sequence without explicit attribution, evaluation defaults to the embedding similarity score alone. A human verification pass is applied to borderline embedding-similarity cases to confirm match validity.

\subsection{Baselines}

We compare X-SYNTH against the following baselines:

\begin{itemize}
  \item \textbf{Small LLM (direct).} A smaller language model provided the raw interaction sequence directly, prompted to identify and synthesize business opportunities, without additional scaffolding.
  \item \textbf{Frontier LLM (direct).} Claude Opus 4.6, prompted identically to the small LLM baseline, serving as a strong upper bound on unstructured LLM capability over the full interaction context.
  \item \textbf{X-SYNTH + Small LLM.} Our proposed agentic framework paired with a smaller underlying model.
  \item \textbf{X-SYNTH + Frontier LLM.} X-SYNTH paired with Claude Opus 4.6 as the reasoning backbone.
\end{itemize}

This design isolates the contribution of the X-SYNTH framework from raw model capability, and characterizes the performance-cost tradeoff across model scales. Beyond performance, we report inference cost across all configurations --- measured in tokens processed and estimated API cost per 125 seller journeys --- to assess the economic viability of each approach at enterprise scale. We defer detailed cost and performance analysis to \S\ref{sec:results}.

\subsection{Evaluation Protocol}

For each benchmark instance, the system receives the full sequence of preceding digital interactions as input and is tasked with producing zero or more proposed business opportunity recommendations. No system has access to the held-out ground-truth filing during inference. Evaluation is conducted in a fully automated pipeline, with human spot-checks applied to ambiguous similarity scores as described above.

\section{Results}
\label{sec:results}

\subsection{Quantitative Results}

Table~\ref{tab:results} summarizes performance across all four configurations on the 302-instance benchmark. X-SYNTH improves both small and frontier models substantially: paired with GPT-4o-mini it raises TLR from 0\% to 57.1\% while reducing false leads from 40 to 6; paired with Claude Opus~4.6 it raises TLR from 9.5\% to 61.9\% (a 6.5$\times$ improvement) while reducing false leads from 50 to 30. The framework contribution is consistent across model scales --- the pattern is not an artifact of raw model capability.

\begin{table*}[htbp]
\small
\begin{tabular}{lrrrrr}
\toprule
\textbf{System} & \textbf{True} & \textbf{Missed} & \textbf{False} & \textbf{TLR} & \textbf{MLR} \\
\midrule
Small LLM (direct)          &   0 & 210 & 40 &  0.0\% & 100.0\% \\
X-SYNTH + Small LLM         & 120 &  90 &  6 & 57.1\% &  42.9\% \\
\midrule
Frontier LLM (direct)       &  20 & 190 & 50 &  9.5\% &  90.5\% \\
X-SYNTH + Frontier LLM      & 130 &  80 & 30 & 61.9\% &  38.1\% \\
\bottomrule
\end{tabular}
\caption{Quantitative results on the 302-instance benchmark (210 positive, 92 negative). True/Missed/False are raw counts; TLR and MLR are computed over 210 positive instances. Small LLM = GPT-4o-mini; Frontier LLM = Claude Opus~4.6.}
\label{tab:results}
\end{table*}

\begin{figure}[htbp]
\centering
\includegraphics[width=\columnwidth]{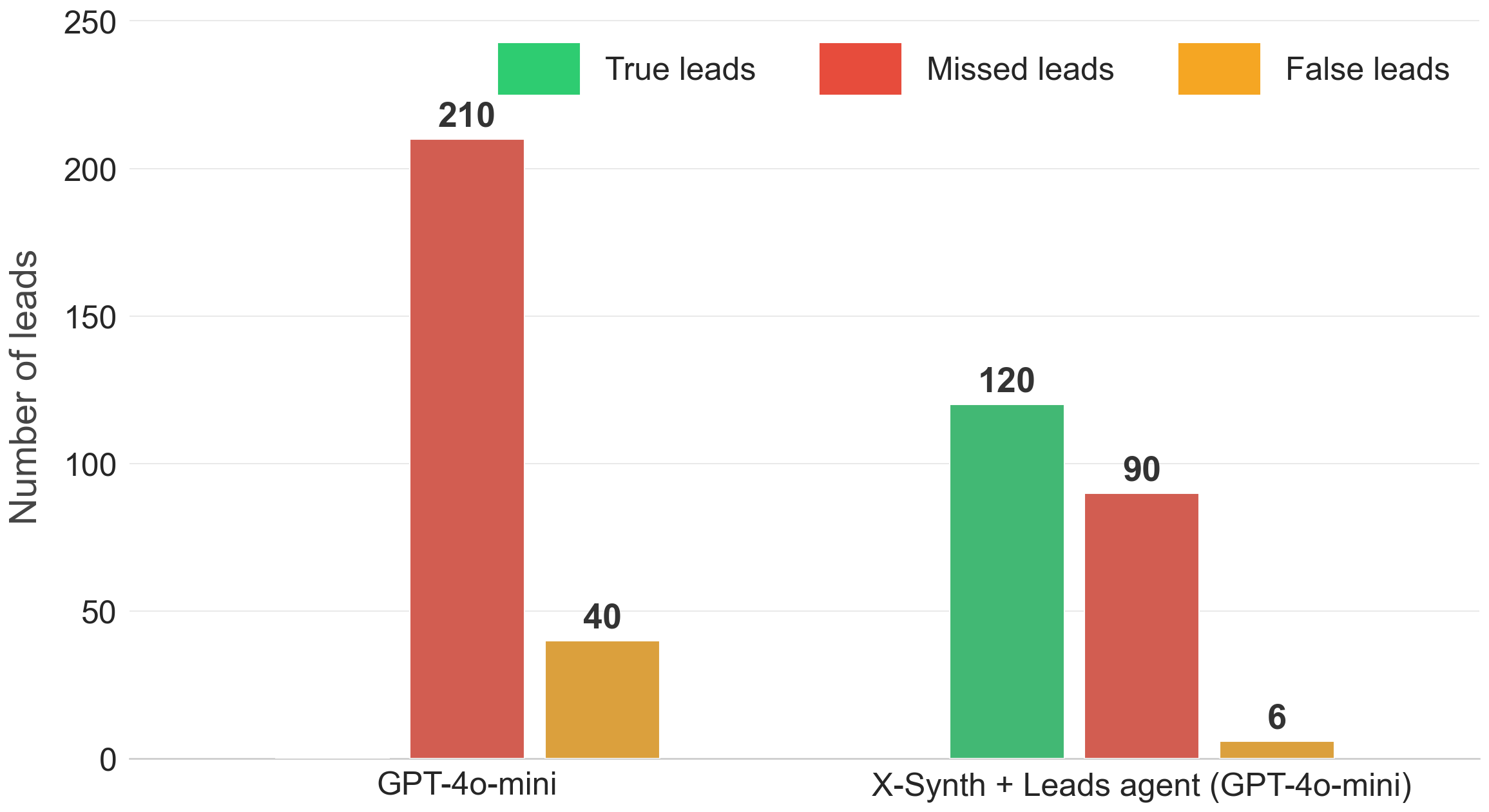}
\caption{\textbf{Lead automation performance across model configurations (small LLM).} X-SYNTH + GPT-4o-mini vs.\ GPT-4o-mini direct. True leads rise from 0 to 120; missed leads fall from 210 to 90; false leads fall from 40 to 6.}
\label{fig:bar-small}
\end{figure}

\begin{figure}[htbp]
\centering
\includegraphics[width=\columnwidth]{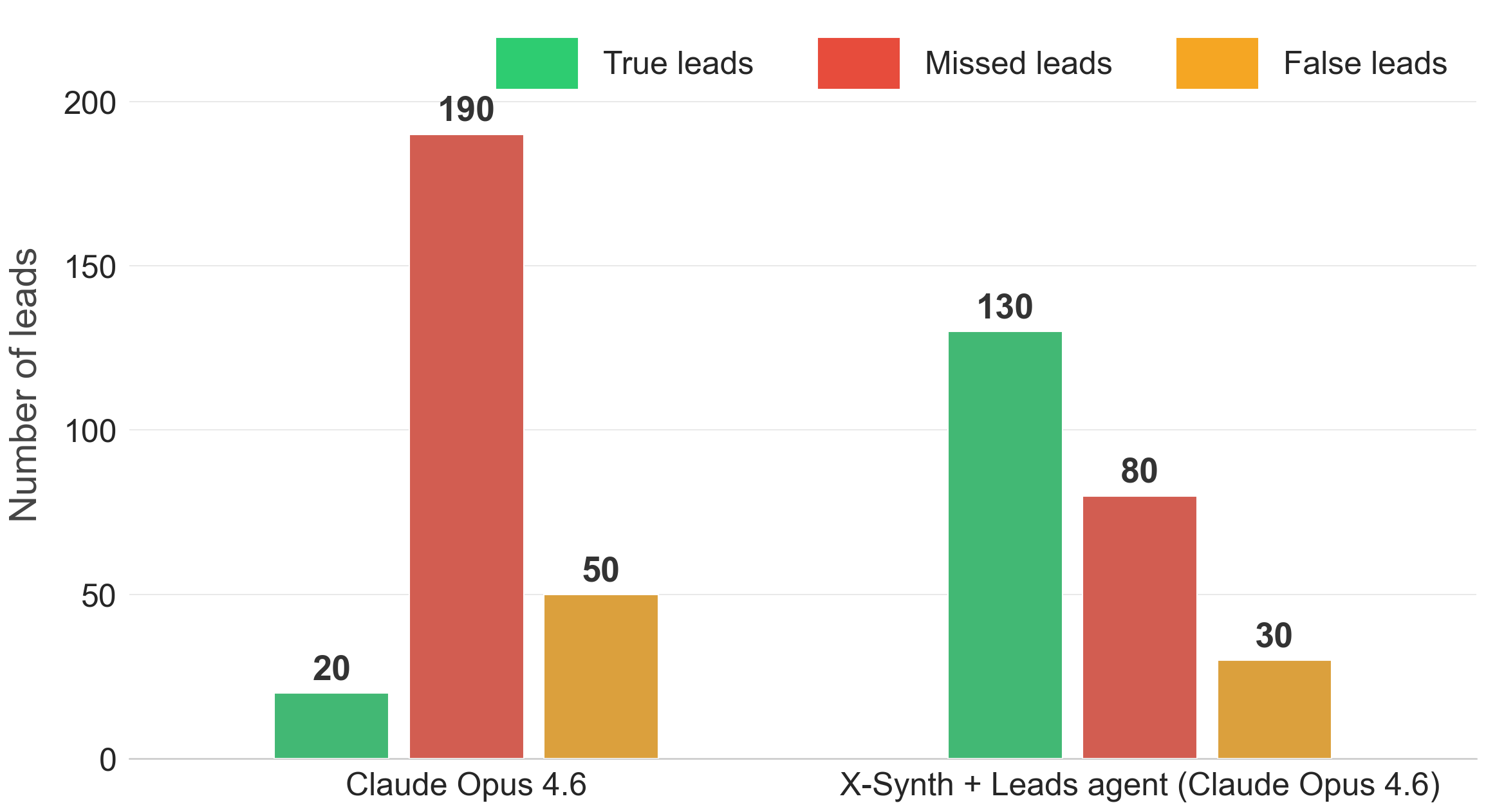}
\caption{\textbf{Lead automation performance across model configurations (frontier LLM).} X-SYNTH + Claude Opus~4.6 vs.\ Claude Opus~4.6 direct. True leads rise from 20 to 130 (6.5$\times$); missed leads fall from 190 to 80; false leads fall from 50 to 30.}
\label{fig:bar-frontier}
\end{figure}

Figures~\ref{fig:bar-small} and~\ref{fig:bar-frontier} show the same results visually. In the small-LLM case, the unaided model finds zero true leads --- the task is entirely outside its reach without structured behavioral context. X-SYNTH converts it into a functional system. In the frontier-LLM case, the baseline is non-zero but still misses 190 of 210 real opportunities; X-SYNTH recovers 110 of those.

\subsection{Qualitative Analysis: The Enterprise Vocabulary Gap}

\begin{figure*}[htbp]
\centering
\definecolor{cardH}{RGB}{26,37,51}
\definecolor{cardB}{RGB}{218,231,244}
\definecolor{cardTS}{RGB}{155,180,205}
\begin{tikzpicture}[
  font=\footnotesize,
  H/.style={fill=cardH, text=white, text width=7.0cm, inner sep=4pt,
            align=left, rounded corners=2pt, minimum height=0.58cm},
  B/.style={fill=cardB, text width=7.0cm, inner sep=4pt, align=left},
]


\node[H, anchor=north west] (h1) at (0,0)
  {\textbf{outlook: opened}\\[1pt]\textcolor{cardTS}{\scriptsize 17:30, 12/3/26}};
\node[B, anchor=north west] (b1) at (h1.south west)
  {Subject: ``TRV QE forecast''\\$\rightarrow$ appointment created};

\node[H, anchor=north west] (h2) at ([yshift=-0.22cm]b1.south west)
  {\textbf{outlook: read}\\[1pt]\textcolor{cardTS}{\scriptsize 17:31, 12/3/26}};
\node[B, anchor=north west] (b2) at (h2.south west)
  {From: sarah.chen@acme.corp\\To: kev@acme.corp\\Subject: updates?\\Body: ``What is going on about XGS Private Ltd?''};

\node[H, anchor=north west] (h3) at ([yshift=-0.22cm]b2.south west)
  {\textbf{CRM: open}\\[1pt]\textcolor{cardTS}{\scriptsize 17:32, 12/3/26}};
\node[B, anchor=north west] (b3) at (h3.south west)
  {$\langle$homepage$\rangle$};

\node[H, anchor=north west] (h4) at ([yshift=-0.22cm]b3.south west)
  {\textbf{CRM: search}\\[1pt]\textcolor{cardTS}{\scriptsize 17:33, 12/3/26}};
\node[B, anchor=north west] (b4) at (h4.south west)
  {Searched: ``XGS Private Ltd''};

\node[H, anchor=north west] (h5) at ([yshift=-0.22cm]b4.south west)
  {\textbf{CRM: view}\\[1pt]\textcolor{cardTS}{\scriptsize 17:34, 12/3/26}};
\node[B, anchor=north west] (b5) at (h5.south west)
  {Viewed: XGS Private Ltd $|$ Opportunities\\Net TCV: 16k\$~$\cdot$~Primary lead: Chandra};


\node[H, anchor=north west] (h6) at (8.5cm,0)
  {\textbf{outlook: view}\\[1pt]\textcolor{cardTS}{\scriptsize 17:36, 12/3/26}};
\node[B, anchor=north west] (b6) at (h6.south west)
  {Meeting reminder: Q2 forecast sync with leadership};

\node[H, anchor=north west] (h7) at ([yshift=-0.22cm]b6.south west)
  {\textbf{outlook: close}\\[1pt]\textcolor{cardTS}{\scriptsize 17:37, 12/3/26}};
\node[B, anchor=north west] (b7) at (h7.south west)
  {Meeting reminder: Q2 forecast sync with leadership};

\node[H, anchor=north west] (h8) at ([yshift=-0.22cm]b7.south west)
  {\textbf{outlook: read}\\[1pt]\textcolor{cardTS}{\scriptsize 17:38, 12/3/26}};
\node[B, anchor=north west] (b8) at (h8.south west)
  {From: kumar@acme.corp\\To: kev@acme.corp\\Subject: Create FZ for PO\\Body: PFA details\\Attachment: CP01000302402.pdf};
\node[draw=orange, line width=1.5pt, rounded corners=3pt, fit=(h8)(b8), inner sep=2pt] {};

\node[H, anchor=north west] (h9) at ([yshift=-0.22cm]b8.south west)
  {\textbf{outlook: write}\\[1pt]\textcolor{cardTS}{\scriptsize 17:43, 12/3/26}};
\node[B, anchor=north west] (b9) at (h9.south west)
  {From: kev@acme.corp\\To: sarah.chen@acme.corp\\Subject: updates?\\Body: XGS Private is currently at a net TCV of 16k\$.\\I'll follow up with Chandra to see if any new RFPs are out.};

\end{tikzpicture}
\caption{Interaction trace for the XGS Private Ltd example. Nine consecutive interaction events are shown in chronological order. The pivotal event (orange border, 17:38) is an email from kumar@acme.corp with subject ``Create FZ for PO'' --- two abbreviations whose meaning is opaque without organizational context.}
\label{fig:xgs-trace}
\end{figure*}
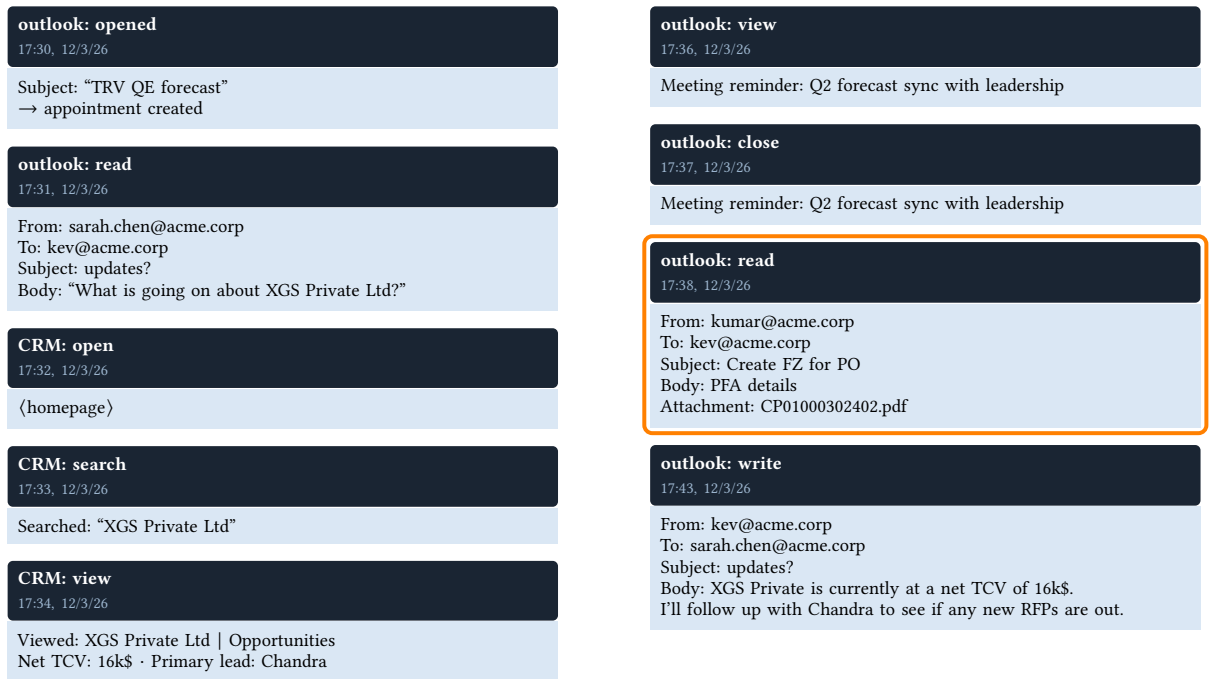

The quantitative gap between baseline and X-SYNTH-paired models is driven in large part by enterprise-specific vocabulary that no general training corpus encodes. Figure~\ref{fig:xgs-trace} shows a representative trace for account XGS Private Ltd.

At 17:38, seller Kev receives an email with subject ``Create FZ for PO'' attaching a contract PDF. Both abbreviations --- FZ and PO --- are organizational terms whose meaning cannot be inferred from their surface form.

\begin{figure}[htbp]
\centering
\includegraphics[width=\columnwidth]{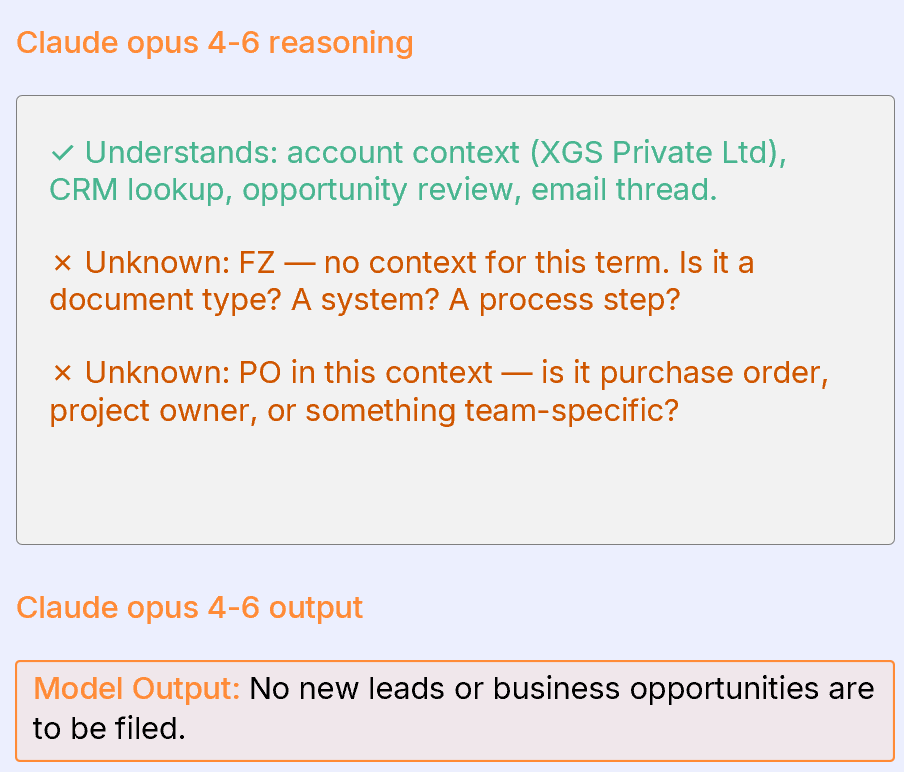}
\caption{Claude Opus~4.6 reasoning: account context and email thread are understood, but FZ and PO are unknown. Output: no new leads.}
\label{fig:xgs-claude}
\end{figure}

Figure~\ref{fig:xgs-claude} shows Claude Opus~4.6's reasoning: it correctly understands the account context, the CRM lookup, and the email thread, but flags FZ and PO as unknown (``Is it a document type? A system? A process step?''). Unable to interpret the request, it outputs: \textit{No new leads or business opportunities are to be filed.}

\begin{figure}[htbp]
\centering
\includegraphics[width=\columnwidth]{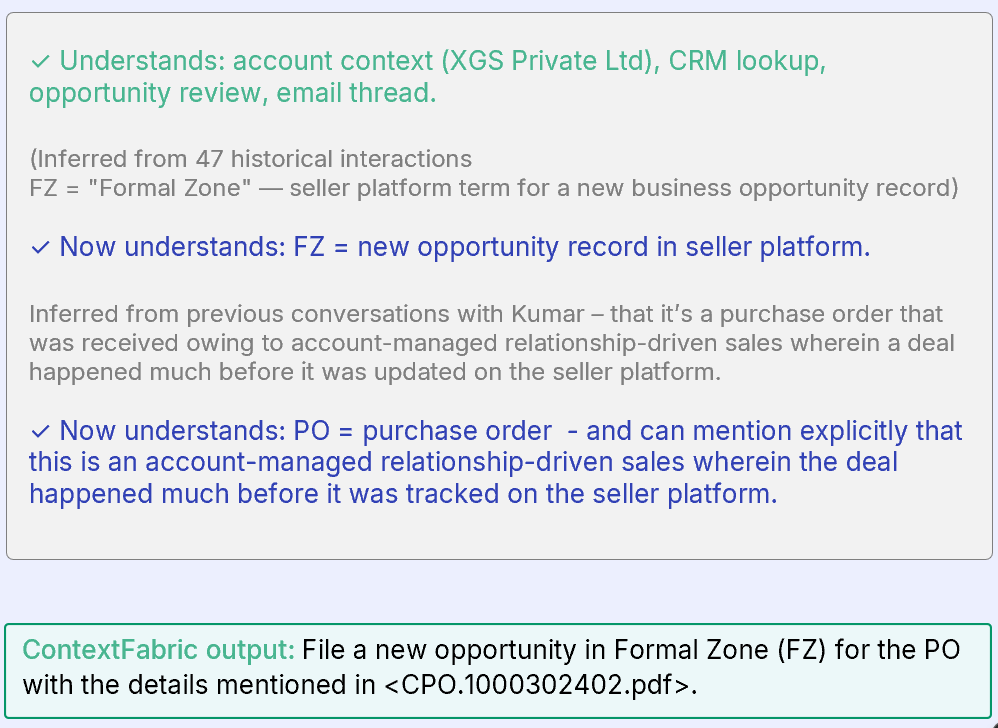}
\caption{X-SYNTH reasoning: FZ inferred as ``Formal Zone'' from 47 prior interactions; PO inferred as purchase order from prior communications with Kumar. Output: file a new opportunity in Formal Zone.}
\label{fig:xgs-xsynth}
\end{figure}

Figure~\ref{fig:xgs-xsynth} shows X-SYNTH's reasoning on the same trace. From 47 prior interactions in the seller's Digital Twin Signature, X-SYNTH infers that FZ = ``Formal Zone,'' the seller-platform term for filing a new business opportunity record. From prior communications with Kumar, it infers that PO = purchase order, and that the deal closed through an account-managed relationship-driven sale before it was tracked on the platform. X-SYNTH outputs: \textit{File a new opportunity in Formal Zone (FZ) for the PO with the details mentioned in the attached contract.}

\subsection{Scale: Why Direct Inference Fails}

\begin{figure*}[htbp]
\centering
\includegraphics[width=\textwidth]{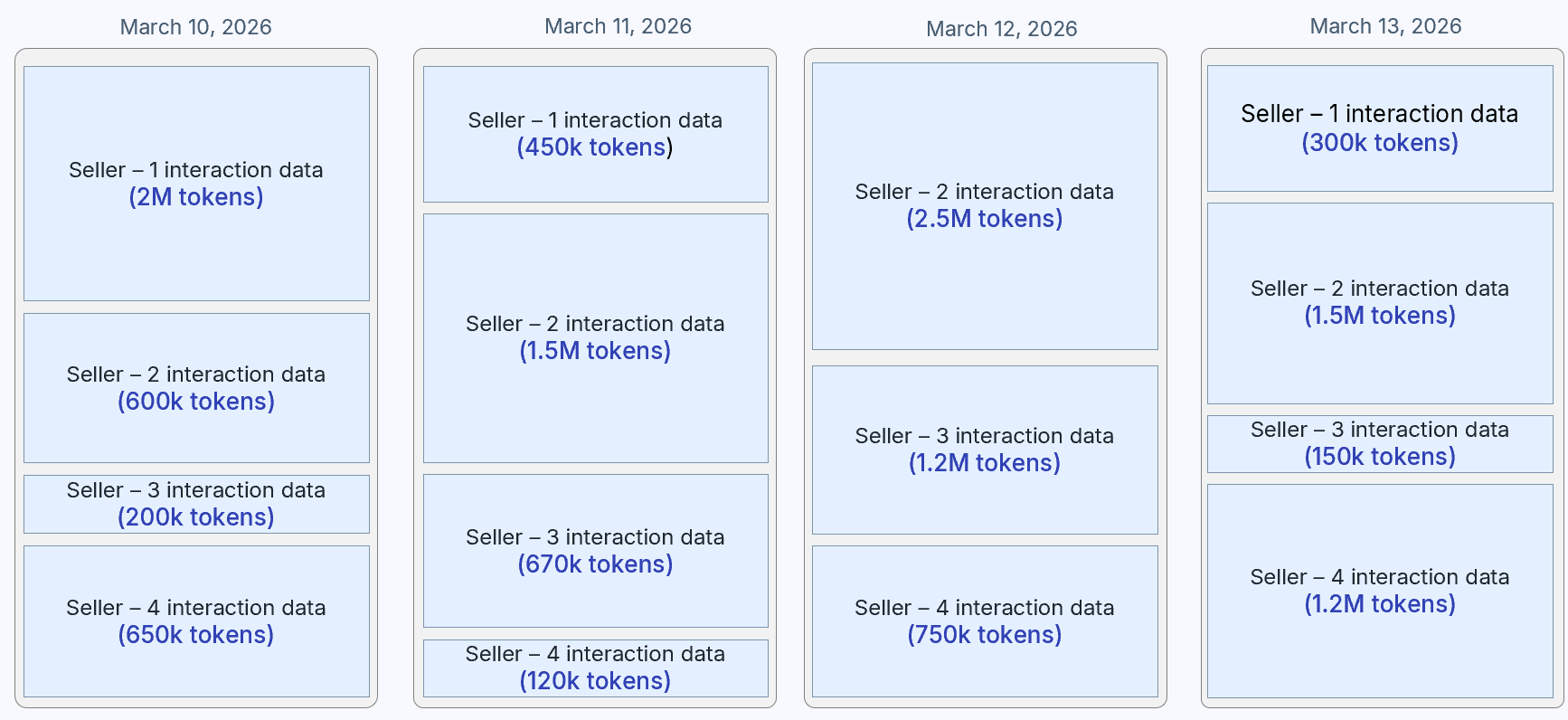}
\caption{Token volume per seller per day across a four-day window (March 10--13, 2026). Individual daily volumes range from 120k to 2.5M tokens; direct inference over the full interaction stream is not economically or technically feasible without selective retrieval.}
\label{fig:token-volume}
\end{figure*}

Figure~\ref{fig:token-volume} illustrates a structural constraint on direct inference: the token volume a single seller generates per day ranges from 120k to 2.5M tokens across a representative four-day window. Passing a full day's interaction stream to a frontier LLM is not feasible at enterprise scale --- the cost is prohibitive and the volume routinely exceeds practical context limits. This is not an edge case; it is the default operating condition for knowledge workers whose role spans multiple accounts, systems, and communication channels simultaneously. X-SYNTH's attention-weighted selective retrieval (Stage~3) is what makes evaluation over real interaction data tractable: rather than ingesting the full stream, it selects the top-$K$ artifacts per individual whose combined attention signal and content relevance are highest for the query.

\subsection{Multi-Day Context Integration}

The third failure mode of direct inference is temporal: a single-day context window misses behavioral signals that only become interpretable across multiple days. Tables~\ref{tab:snippets-single} and~\ref{tab:snippets-multi} show a paired example for seller s.chen and account AC (Acme Corp).

Table~\ref{tab:snippets-single} shows the five March~13 interactions that Claude Opus~4.6 receives as input. The model surfaces a Globex renewal follow-up (a genuine signal from the Slack message at 15:10) but produces no recommendation for an Acme Corp streaming opportunity. The interactions are coherent within the single day, but the signal is too weak: one Helix ticket, one Vault document review, and one Slack message do not, in isolation, establish a clear opportunity.

\begin{table*}[htbp]
\footnotesize
\begin{tabular}{p{2.3cm}p{1.0cm}p{1.0cm}p{1.8cm}p{9.6cm}}
\toprule
\textbf{Timestamp} & \textbf{App} & \textbf{User} & \textbf{Action} & \textbf{Screen Text} \\
\midrule
2026-03-13 07:45 & Gmail & s.chen & read &
From: D.Kim --- Should I pull any background docs before the AC call or are we going in light? Also --- are we doing a follow-up with the Globex team after their renewal or is that on you? \\
\addlinespace
2026-03-13 08:10 & Helix & s.chen & viewed ticket &
2026-09213: event delivery timing inconsistent with expectations; latency outside acceptable range. Opened: 31 days. Owner: unassigned. Account: AC. Impacted rev: not captured. \\
\addlinespace
2026-03-13 08:33 & Vault & s.chen & viewed document &
AC MSA v2.1. Sections viewed: 7 (9 min), 8 (2 min). Last viewed by: s.chen. \\
\addlinespace
2026-03-13 14:02 & Zoom & s.chen & joined meeting &
AC Q1 review. Attendees: s.chen, d.kim, 2 external. Duration: 44 min. Recording: off. Transcript: none. \\
\addlinespace
2026-03-13 15:10 & Slack & s.chen & read message &
\#se-pool --- D.Kim: good session, SC --- sending you that Globex renewal doc. Also flagging: AC streaming module expansion with RiverFlow is in your manager's queue before you decide how to handle it. \\
\bottomrule
\end{tabular}
\caption{A single day of interaction data (March 13) as consumed by Claude Opus~4.6. The model surfaces a Globex renewal follow-up but produces no recommendation for the Acme Corp (AC) streaming opportunity --- the signal is present but the single-day window lacks the behavioral accumulation needed to infer it.}
\label{tab:snippets-single}
\end{table*}

Table~\ref{tab:snippets-multi} shows the ten interaction snippets that X-SYNTH selects across four days (March 10--13). The selection is driven by the DTS-conditioned attention filter: X-SYNTH identifies that s.chen's attention has been accumulating on AC-related artifacts across multiple applications over multiple days, and retrieves the artifacts that account for the most behavioral signal. The result is a lead trace that no single-day window could assemble.

\begin{table*}[htbp]
\footnotesize
\begin{tabular}{p{1.6cm}p{1.0cm}p{1.0cm}p{1.8cm}p{10.3cm}}
\toprule
\textbf{Timestamp} & \textbf{App} & \textbf{User} & \textbf{Action} & \textbf{Screen Text} \\
\midrule
03-10 09:14 & Slack & s.chen & read thread &
JW asking whether platform handles high-frequency event delivery at low latency. D.K: RF has been visible in their org the past few weeks. \\
\addlinespace
03-11 10:30 & Lens & s.chen & ran report &
account: AC. add\_on\_1--3: no license; streaming: no license. active\_users: 2 of 9. \\
\addlinespace
03-11 08:09 & Gmail & s.chen & read &
We have been working around a limitation for some time and wanted to understand whether there is a better path forward or whether what we have is as good as it gets. \\
\addlinespace
03-11 08:22 & Helix & s.chen & searched &
ticket 9203: ingestion behavior reported --- open, unassigned. \\
\addlinespace
03-12 08:02 & Vault & s.chen & viewed document &
AC MSA v2.1. Sections: 3 (7 min), 7 (3 min). Last amended: never. \\
\addlinespace
03-12 08:44 & Helix & s.chen & searched &
query: add-on pricing tier 2 accounts FSI. result\_1: list price range by tier FSI. result\_2: discount bracket FSI accounts. \\
\addlinespace
03-12 10:11 & Slack & s.chen & sent message &
Kafka environment, FSI context --- RF may be relevant background. D.Kim: yes, treating as technical standby for now. \\
\addlinespace
03-12 08:02 & Vault & s.chen & viewed document &
AC MSA v2.1. Sections: 7 (6 min), 8 (2 min). Last amended: never. \\
\addlinespace
03-13 08:33 & Vault & s.chen & viewed document &
AC MSA v2.1. Sections: 7 (9 min), 8 (2 min). Last viewed by: s.chen. \\
\addlinespace
03-13 15:10 & Slack & s.chen & read message &
D.Kim: flagging the RF thing we discussed, worth looping in your manager before you decide how to handle it. \\
\bottomrule
\end{tabular}
\caption{X-SYNTH pulls 10 interaction snippets across four days (March 10--13) from Gmail, Slack, Helix, Lens, Vault, and Meridian --- each weak or ambiguous alone --- and integrates them into a single lead trace. The trace establishes: AC has no streaming module license and is actively working around the gap; RiverFlow is pitching into the same gap; s.chen has completed discovery, reviewed contract sections, and sized the deal --- but has not filed. X-SYNTH recommends: file a new streaming module expansion opportunity for AC, flagged as active competition.}
\label{tab:snippets-multi}
\end{table*}

The five March~13 interactions in Table~\ref{tab:snippets-single} are a strict subset of the ten interactions in Table~\ref{tab:snippets-multi}. Claude Opus~4.6 had access to the same March~13 data and missed the opportunity; X-SYNTH extended the retrieval window to four days, surfacing five additional artifacts that together made the AC streaming gap, RiverFlow's presence, and the seller's accumulated deal activity legible as a filing-ready opportunity. The miss was not caused by insufficient raw data --- it was caused by the absence of a mechanism to select the right temporal window.

\subsection{Summary: What Drives the Improvement}

The quantitative gap between X-SYNTH and direct inference is not attributable to a single factor. Three structural properties of enterprise knowledge work, each invisible to a model operating over raw interaction streams, account for the improvement collectively.

\textbf{Enterprise context and priors.} General-purpose models lack the organizational priors needed to interpret enterprise signals correctly. The XGS Private Ltd example illustrates the failure mode: two opaque abbreviations are sufficient to suppress an otherwise detectable opportunity. Correct interpretation requires priors built from the seller's own prior interaction history --- priors that must be established before inference, because each new interaction implicitly depends on what preceded it. X-SYNTH grounds inference in those accumulated priors rather than in model training data alone.

\textbf{Scale and selective retrieval.} A single seller generates up to 2.5M tokens of interaction data per day --- a volume that makes direct inference over the full stream economically and technically infeasible. X-SYNTH's attention-weighted retrieval collapses this to a tractable, high-signal subset without requiring manual filtering.

\textbf{Cross-temporal and cross-individual integration.} Real sales opportunities accumulate as behavioral signals across days and across team members, not within a single session or a single actor's view. The Acme Corp example shows that five individually ambiguous March~13 interactions become a filing-ready lead trace only when extended with artifacts from the preceding three days and from collaborators who share the account context. No single-day, single-individual context window can recover this signal.

These three properties are not independent failure modes --- they compound. An opportunity that requires cross-day, cross-individual context, involves organizational vocabulary, and sits within a high-volume interaction stream will be missed by direct inference for all three reasons simultaneously. X-SYNTH addresses each structurally, which is why the improvement is consistent across both model scales tested.

The common enabler across all three properties is the DTS-conditioned attention filter. Rather than applying a fixed retrieval strategy uniformly, X-SYNTH conditions filter selection jointly on the query and each individual's Digital Twin Signature --- a rolling behavioral profile encoding domain attention, rhythm, responsibility, and recent divergence from their own baseline. This single architectural decision is what makes each of the three failure modes recoverable:

\textbf{Enterprise context and priors.} A general model has no organizational memory. The DTS does. Because it is built continuously from the seller's own interaction history, it encodes the vocabulary, relationships, and conventions specific to that individual and that organization --- and it encodes them before any new query arrives. When a new interaction is observed, the DTS-conditioned filter interprets it against that accumulated context rather than against generic training priors. The organizational knowledge is not retrieved; it is already embedded in the relevance signal.

\textbf{Scale and selective retrieval.} Without a behavioral prior, a retrieval system has no principled basis for distinguishing a high-signal artifact from noise at enterprise token volumes. The DTS provides that prior: the Proportional, Differential, and Recurrent filters each define importance relative to what is normal for this individual, which means the top-$K$ selection is not keyword matching against the full stream but a behaviorally-grounded compression that discards noise by definition.

\textbf{Cross-temporal and cross-individual integration.} A single-day, single-actor window misses signals that only become legible across time and across a team. The DTS directly encodes this structure: the short-vs-long divergence component flags when recent behavior departs from the two-week baseline, making multi-day accumulation observable; the Collective filter aggregates attention across the scoped cohort $\mathcal{U}_q$, making cross-individual signals --- consensus focus, individual outliers --- first-class inputs to retrieval. The temporal and team-level context is not reconstructed after the fact; it is baked into the relevance computation.

The improvement is not additive across three separate fixes. It is the consequence of replacing a query-only relevance function with one that is conditioned on who is doing the work, what their behavioral history looks like, and how that history situates the current moment.

\section{Conclusion}
\label{sec:conclusion}

Enterprise AI agents are bottlenecked not by reasoning capability but by the absence of a principled relevance signal. This paper presents X-SYNTH, a framework that grounds enterprise context synthesis in observed digital human attention. Digital human attention, as operationalized here, is the ordered, digitally observable interaction signature of each worker --- encoding not just what was done, but the sequence in which it was done and the implicit reward signals embedded within. Each individual's behavioral baseline is modeled as a Digital Twin Signature (DTS), a compact rolling behavioral profile built from five components that together characterize recency, domain focus, diversity, and sequential structure. A learned router selects among seven qualitatively distinct attention filters --- Proportional, Inverse, Differential, Recurrent, Comparative, Sequential, and Collective --- per individual and per query, conditioned jointly on the query embedding and the DTS. A four-stage agentic pipeline assembles the result: subject scoping, individualized modality selection, attention-and-content-weighted retrieval, and synthesis with credit-attributed feedback. On a 302-instance benchmark drawn from real Fortune 500 enterprise interaction data, augmenting Claude Opus~4.6 with X-SYNTH raises True Lead Rate from 9.5\% to 61.9\% (a 6.5$\times$ improvement) while reducing False Lead Rate from 90.5\% to 18.8\%, surfacing 110 additional real leads that the unaugmented model misses entirely. Enterprise context synthesis is not a retrieval problem. It is a relevance problem --- and digital human attention is its most reliable ground truth.

\bibliographystyle{ACM-Reference-Format}
\bibliography{references}

\end{document}